\documentclass[5p,times]{elsarticle}
\usepackage{amsmath}
\usepackage{booktabs}

\usepackage{amssymb}

\usepackage{microtype}
\usepackage{fontawesome5}
\usepackage{url}

\usepackage[bookmarks=false]{hyperref}
    \hypersetup{colorlinks,
      linkcolor=blue,
      citecolor=blue,
      urlcolor=blue}

\begin{document}
\begin{frontmatter}

\title{RULix:\texorpdfstring{\\}{ }A Synthetically Pretrained Foundation Model for Remaining Useful Life Prediction}

\author[a]{Felix Saretzky\corref{cor1}\fnref{eq}}
\author[b,c]{Andreas Pauly\fnref{eq}}
\author[a]{Thomas Engel}
\author[b,c]{Jakob Rehof}

\address[a]{Department of Computer Science, SECAN-Lab, University of Luxembourg, Luxembourg}
\address[b]{Department of Computer Science, TU Dortmund University, Dortmund, Germany}
\address[c]{Lamarr Institute, Dortmund, Germany}

\cortext[cor1]{Corresponding author. {\it E-mail address:} felix.saretzky@uni.lu}
\fntext[eq]{These authors contributed equally to this work.}

\begin{abstract}
Maintaining manufacturing machines is costly and hard to schedule: operators must reserve a maintenance window, procure replacement parts, and decide which machine to service first.
Data-driven remaining-useful-life (RUL) estimation predicts when a machine will fail and enables planning ahead, yet it requires run-to-failure histories that production assets rarely provide. 
Preventive maintenance truncates degradation before failure, and industrial fleets are small, so labeled degradation data remain scarce at any single site.
Recent studies therefore repurpose time-series foundation models for prognostics, while pretraining on degradation processes remains largely unexplored.
This paper presents RULix, a prognostic foundation model pretrained exclusively on synthetic run-to-failure data.
A synthetic prior generates sensor data from degradation processes based on stochastic differential equations (SDEs) and observed through structural causal graphs unrolled over time.
An extended long short-term memory (xLSTM) encoder embeds sliding sensor windows in a latent space that represents machine dynamics, from which we predict the machine's health state, degradation dynamics, and RUL.
The pretrained model is evaluated without fine-tuning on nine run-to-failure benchmarks, including the C-MAPSS turbofan benchmarks, and outperforms trained models when only two to four run-to-failure units are available.
The approach is a step toward prognostic foundation models for assets with few recorded failure histories.

\medskip\noindent\faGithub\ Code and weights: \url{https://github.com/FelixSaretzky/RULix}

\end{abstract}


\end{frontmatter}

\section{Introduction}
\label{sec:intro}

No sector spends more on industrial maintenance than manufacturing, since stoppages of automated production lines are very costly~\cite{erkoyuncu2026maintenance}.
Remaining useful life (RUL) estimation predicts when a machine will fail and gives operators the lead time to reserve a maintenance window, procure spare parts, and decide which machine to service first.
A discrete event simulation shows that predictive maintenance outperforms reactive and planned maintenance even for uncertain RUL estimates, yet many companies remain reluctant to adopt it because the investment costs are high and the estimates are uncertain~\cite{benker2022maintenance}.

These barriers weigh most on small and medium-sized enterprises (SMEs).
Compared with large companies, SMEs have smaller budgets, fewer analytics skills and limited access to data, and they therefore benefit from ready-to-use models that can be deployed without much engineering work or specialist knowledge~\cite{kaiser2023sme}.
Implementing machine learning for predictive maintenance requires substantial data preparation and interdisciplinary teams, which often go beyond what SMEs can provide in experience, skills and staff~\cite{welte2020sme}.

Independent of company size, data-driven RUL models learn from run-to-failure histories with RUL labels, which production assets rarely provide.
Most recorded data stem from healthy operation~\cite{li2024smalldata}, and preventive maintenance replaces components before the final stage of degradation is recorded~\cite{xiong2023physics}.
Complete run-to-failure cases are therefore rarely collected in manufacturing~\cite{kim2025sparse}, and most manufacturers cannot afford to collect monitoring data at scale~\cite{li2024smalldata}.

In addition, prognostic models are typically developed for individual components and generalize poorly~\cite{theiler2026tabpfn,nunes2022challenges}, so every new asset, configuration or operating regime requires its own data and engineering effort.
A recent review in the CIRP Annals accordingly counts poor transfer of models between assets, a lack of explainability and an incomplete handling of uncertainty among the main open gaps in digitally enhanced maintenance~\cite{erkoyuncu2026maintenance}.
Two lines of research address the missing data.
Physics-informed machine learning and synthetic data generators add prior knowledge, but this knowledge is tied to the asset under study~\cite{xiong2023physics, braun2026piml,pauly2026pumps}, and the effort of embedding it cannot be amortized across assets~\cite{braun2026piml}.
Foundation models promise one model for many assets, but those applied to RUL estimation so far are pretrained for forecasting or generic tabular tasks and still need labeled failure histories of the target system to learn the mapping from sensor data to RUL~\cite{theiler2026tabpfn, elghoussani2026timeseriesfoundationmodelembeddings}.

This paper presents RULix, a prognostic foundation model pretrained exclusively on synthetic degradation processes, in line with the call for foundation models of degradation that serve as one backbone across assets~\cite{braun2026piml}.
The premise is that degradation shares a generic structure across assets: a latent health state that drifts under noise until it reaches a failure level \cite{jardine_review_2006}.
This structure requires no asset-specific physics model and can be simulated at scale, which makes a synthetic prior a viable substitute for scarce run-to-failure data.
An xLSTM encoder reads a window of sensor data, and two heads infer the latent health state and the drift and diffusion functions of the degradation process.
The RUL follows from simulating the inferred process until it first reaches the failure level.
RULix learns the mapping from sensor data to degradation from synthetic data only and needs no fine-tuning on the target system, which lowers the entry barrier for companies without in-house data science expertise.
At inference, it combines the sensor history of the machine with calibration information from a few past units that ran to failure: a healthy reference and a failure level for scaling, and their lifetimes for the time before degradation sets in.
Such failure times are typically recorded in maintenance logs and maintenance management systems~\cite{erkoyuncu2026maintenance}.
Each prediction comes with an uncertainty interval and decomposes into the probability that degradation has begun, the current health state, the inferred degradation law and the simulated time to failure, which makes it traceable for maintenance planners.

The contributions are: (i) a synthetic prior of run-to-failure data in which a latent health SDE is observed through a temporal structural causal model, (ii) RULix, an xLSTM encoder with health and operator heads that obtains the RUL by first-passage simulation and is calibrated to the target system without fine-tuning, and (iii) an evaluation on real run-to-failure benchmarks that relates the prediction error to the amount of calibration information.

\section{Related Work}

Classical statistical methods as reviewed by \cite{si2010rulstat} consider the degradation as a stochastic process with failure as its first passage.
But their parameters are fitted per application and cannot be transferred.
In \cite{elghoussani2026timeseriesfoundationmodelembeddings} a frozen time-series foundation model is used to encode the data, and a new MLP is trained to predict the RUL. 
This requires labeled run-to-failure data of the target system, which are scarce, whereas RULix is trained on synthetic data and uses past information only at inference.
The closest approach is SurvivalPFN~\cite{qi2026survivalpfn}, which requires a population with labeled event data as context, whereas RULix operates only on the sensor history of one machine and past information. 

\section{Background}
\label{sec:background}
RULix is applied in an industrial setting, where machines are monitored by sensors.
These observations form a multivariate time series $y^{1:t} = (y^1 \dots y^t)$ with $y^i \in \mathbb{R}^d$ sampled at equidistant time steps, like operation cycles.
The number of sensors $d$ might differ for every machine.
Additionally, we assume that we might have expert knowledge about the system, like operating conditions or previous RUL values.
This data might be present in different forms, as intervals, lists or just a single measurement. 

This section introduces the basic components of RULix, SDE to model the degradation, PFN as the inference principle and xLSTM as the causal sequence encoder.

\subsection{Stochastic Differential Equation (SDE)}
\label{sec:bg-sde}
A stochastic differential equation (SDE) describes a con\-tin\-u\-ous-time stochastic process with a deterministic trend and random perturbation.
It could be described as an ordinary differential equation driven by white noise $w(t)$,
\begin{equation}
    \frac{\mathrm{d}x}{\mathrm{d}t} = f(x, t) + L(x, t)\, w(t),
    \label{eq:sde-heuristic}
\end{equation}
with state $x \in \mathbb{R}^n$, where the drift $f(x, t)$ describes the deterministic trend and the dispersion $L(x, t) \in \mathbb{R}^{n \times m}$ determines how the noise influences each component of the state $x$.
White noise is the formal derivative $\mathrm{d} W(t)/\mathrm{d}t$ of a standard Wiener process $W(t) \in \mathbb{R}^m$, which has independent Gaussian increments  $\Delta W_k = W(t) - W(s) \sim \mathcal{N}(0, (t - s)\, I)$ for $t > s$. 
Since $W$ is nowhere differentiable, \eqref{eq:sde-heuristic} is interpreted as the It\^{o} integral
equation~\cite{sarkka2019applied}
\begin{equation}
    x(t) = x(t_0) + \int_{t_0}^{t} f(x(\tau), \tau)\, \mathrm{d}\tau
                  + \int_{t_0}^{t} L(x(\tau), \tau)\, \mathrm{d}W(\tau),
    \label{eq:sde}
\end{equation}
commonly abbreviated as $\mathrm{d}x = f(x, t)\, \mathrm{d}t + L(x, t)\, \mathrm{d}W$.

Which is approximated numerically with the Euler--Maru\-yama method on a grid $t_k = t_0 + k \Delta t$
\begin{equation}
    \hat{x}(t_{k+1}) = \hat{x}(t_k) + f(\hat{x}(t_k), t_k)\, \Delta t
                     + L(\hat{x}(t_k), t_k)\, \Delta W_k,
    \label{eq:euler-maruyama}
\end{equation}
where $\hat{x}$ is the approximation and $\Delta W_k = W(t_{k+1}) - W(t_k)$ are independent increments of the Wiener process.

\subsection{Amortized Inference with Prior-data fitted networks}
\label{sec:bg-pfn}
Prior-data fitted networks (PFNs) \cite{mueller2021pfn} are trained on synthetic data drawn from a prior over data-generating mechanism $\phi \sim p(\phi)$, $D \sim p(D \mid \phi)$.
Minimizing the prior-data negative log-likelihood (NLL)
\begin{equation}
    \ell(\theta) = \mathbb{E}_{D \cup \{(x, y)\} \sim p(D)}\big[-\log q_\theta(y \mid x, D)\big]
    \label{eq:pfn-loss}
\end{equation}
trains $q_\theta$ to approximate the posterior predictive distribution (PPD)
\begin{equation}
    p(y \mid x, D) = \int_{\Phi} p(y \mid x, \phi)\, p(\phi \mid D)\, \mathrm{d}\phi ,
    \label{eq:ppd}
\end{equation}
which the trained network returns for a new data set $D$ in a single forward pass,
without gradient updates (in-context learning).
The approximation holds under the prior, so the generator has to cover the task at hand. 

If the target of the PFN is not observed in $D$ but determined by $\phi$, such as a function $f_\phi(x)$ which is evaluated at $x$, the objective is not changed and $q_\theta$ approximates the posterior $p(f_\phi(x) \mid D)$ instead of the PPD.
This is amortized inference \cite{Gershman2014AmortizedII}, where the cost of inference is paid once during the training and reused for every new data set.

The foundation inference model for SDEs (FIM-SDE) \cite{seifner2025incontext} applies this idea to It\^{o} SDEs.
Given noisy, irregular sampled paths of a dynamical system as context, it estimates drift and diffusion at query locations in the region covered by observations.
Instead of using the network to predict the path, it infers these functions and simulates paths with Euler--Maruyama from initial states.

\subsection{Extended Long Short-Term Memory}
\label{sec:bg-xlstm}

Extended Long Short-Term Memory (xLSTM) \cite{beck2024xlstm} is an updated version of the LSTM that introduces exponential gating, normalization, stabilization and matrix memory.
xLSTM introduces two specific layers for that, mLSTM and sLSTM.
Both layer types use exponential gating to update the memory with incoming values.
sLSTM mixes the various memory cells by using recurrent connections, which happens stepwise.
mLSTM replaces scalar values by a matrix memory whose cells are not mixed, so it can be computed in parallel. 
It is updated with a covariance update rule and with its gates it implements a parallel form of linear attention. 
The xLSTM model is formed by stacking sLSTM and mLSTM in residual blocks. 

PFN as introduced in \cite{mueller2021pfn} assumes that examples in $D$ are i.i.d.\ given $\phi$ $p(D \mid \phi) = \prod_{i} p(x_i, y_i \mid \phi)$. 
But a time series is ordered and the likelihood factorizes causally over the sequence  $p(y_{1:t} \mid \phi) = \prod_{s=1}^{t} p(y_s \mid y_{1:s-1}, \phi)$.
xLSTM is capable of processing the sequence step by step, with a target for each time step during the training and only knows the past, such that it can predict the posterior distribution for every step.

\section{RULix}
\label{sec:method}

\begin{figure*}[t]
\centering
\includegraphics[width=\textwidth]{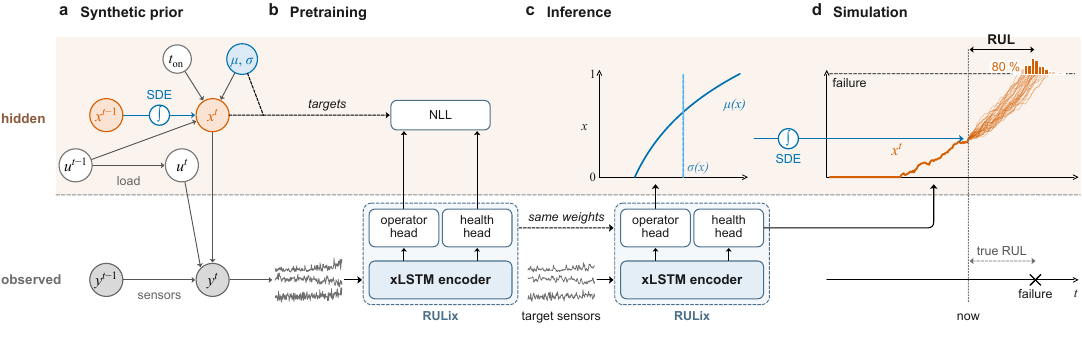}
\caption{Overview of RULix. (a) Synthetic prior: after the onset $t_{\mathrm{on}}$, the degradation $x^t$ follows an SDE
 driven by the load $u^t$, and a temporal SCM emits the sensors $y^t$. Colored nodes are the training targets. (b)
 Pretraining of the xLSTM encoder and its heads against these targets, without RUL labels. (c) Inference on a target unit with the same weights. (d) RUL from SDE paths with the inferred operators to the failure level $x = 1$, median
 and 80\,\% interval.}
\label{fig:overview}
\end{figure*}

RULix infers a machine's health state and degradation law from a window of sensor readings and obtains the remaining useful life by simulating the inferred dynamics to failure.
Figure~\ref{fig:overview} gives an overview.
We detail the four steps in the following subsections.

\subsection{Problem statement}
\label{sec:setting}

Consider an industrial machine monitored with $d$ sensors, where some sensors measure operating conditions of the machine, such as the load, this produces a sensor history $y^{1:t}$ with $y^s \in \mathbb{R}^d$.
This machine fails at time $T$, the failure might differ between datasets e.g., breakdown or that the capacity of a battery is below a threshold. 
Remaining useful life (RUL) is a conditional random variable $R_t = T - t \mid T > t, y^{1:t}$ \cite{jardine_review_2006}.
Our objective is to approximate its distribution given the sensor history $p(R_t \mid y^{1:t})$ with a model $q_\theta (R_t \mid y^{1:t})$.
As our assumption is that run-to-failure data is scarce, we solely use synthetic data to learn $q_\theta$. 
The trained model can be applied zero-shot, from the pure sensor history, or in an informed way that uses calibration information $\mathcal{C}$ about machine type, such as lifetimes of similar machines and their failure level, from a fleet or experts.
The prediction then becomes $q_\theta (R_t \mid y^{1:t}, \mathcal{C})$, $\mathcal{C}$ is used only in inference and not during training.
In our experiments, $\mathcal{C}$ is always given as the run-to-failure training units of
each dataset.

\subsection{Data generation -- Prior}
\label{sec:prior}

Each synthetic unit is drawn with its own mechanism $\phi$.
As operating condition we insert load $u(t)$ that follows an AR(1) process around one.
The latent degradation $x(t) \in \mathbb{R}$ stays at zero until the onset and then follows a scalar SDE as in Sec.~\ref{sec:bg-sde} with $f(x, t) = u(t)\,\mu(x)$ and $L(x, t) = \sigma(x)$
\begin{equation}
  dx(t) = u(t)\, \mu\big(x(t)\big)\, dt + \sigma\big(x(t)\big)\, dW(t),
\end{equation}
with drift $\mu = \mu_0 \psi_\mu$ and diffusion $\sigma = \sigma_0 \psi_\sigma$, built from base rates $\mu_0, \sigma_0$ and shapes $\psi_\mu(0) = \psi_\sigma(0) = 1$ from a library of degradation laws.
We simulate it with Euler--Maruyama \eqref{eq:euler-maruyama} at the sampling steps ($\Delta_t = 1$),
\begin{equation}
  x^{t+1} = x^t + u^t \mu(x^t) + \sigma(x^t)\, \Delta W^t,
  \qquad \Delta W^t \sim \mathcal{N}(0, 1),
\end{equation}
and assume that the unit fails at $T = \min\{t \mid x^t \ge 1\}$ \cite{si2010rulstat}. 
The prior simulates $d$ sensors, from this set of sensors two of them observe the load with noise and the others form a temporal structural causal model on a random graph, 
\begin{equation}
  y^t_i = h_i\Big(\textstyle\sum_j G_{ij}\, y^t_j + a_i\, y^{t-1}_i
          + A_i\, r_i(x^t) + l_i\, u^t\Big) + \beta_i\, \varepsilon^t_i ,
\end{equation}
with instantaneous edges $G$, autoregression $a_i$, nonlinearity $h_i$, degradation response $r_i$ (linear, power or hinge), load coupling $l_i$ and noise $\varepsilon^t_i$ with scale $\beta_i$. 
To make the degradation readable from the sensors, we calibrate $A$ such that the noise-free sensor excursion between $x = 0$ and $x = 1$ has unit norm, scaled by a per-unit factor with median one.
The training targets are the degradation $x^{1:T}$ and the drift and diffusion functions $\mu$ and $\sigma$ evaluated on a grid.

\subsection{RULix -- Model and Inference}
\label{sec:model}

We now introduce RULix, an xLSTM-based inference model based on ideas from FIM-SDE \cite{seifner2025incontext}.
From a window of sensor history it estimates the current degradation state and the drift and diffusion of the associated SDE and obtains the RUL by simulation. 

Every unit has different measures and time scales.
Therefore we apply causal scaling of the sensors based on the data seen so far, these statistics are passed to the model as an additional input.
Time scales are not scaled, we assume that every measurement is one operating cycle and that the prior covers a reasonable amount of applications.

A causal xLSTM encoder with three mLSTM blocks and one sLSTM block maps the window of sensor data to one hidden state per step. 
The health head is a linear layer that predicts per step a zero-inflated Gaussian over $x^t$, the probability that the unit is still healthy and a Gaussian otherwise.
The operator head follows the attention operator of FIM.
A fixed grid of 65 health values in $[0, 1]$ queries the encoded window by cross-attention and returns a Gaussian over $\log\mu(x)$ and $\log\sigma(x)$ at every grid point.

The loss is the sum of health NLL and operator NLL. 
The health NLL covers all steps of the window and is weighted exponentially towards the query step.
The operators are supervised in standardized log space. 

To predict RUL at step $t$ as zero-shot, we draw starting states from the health posterior and simulate the SDE with predicted operators with Euler--Maruyama at constant load $u=1$ until the first passage of $x = 1$ such that the number of integration steps is a sample of $R_t$ for that path. 
A path that starts healthy enters the degradation ramp immediately.
Repeated over 256 paths, the first-passage times form $q_\theta(R_t \mid y^{1:t})$, of which the median and the 10--90\,\% interval are reported.

Where the failure threshold lies on the sensor scale cannot be determined from the sensor history alone. 
We therefore use calibration information $\mathcal{C}$, in our experiments the run-to-failure training units as a surrogate for expert knowledge: the median sensor excursion between their healthy start and their end of life sets the failure threshold $x = 1$, their RUL labels give the duration of one model step, healthy paths instead draw their RUL from the residual lives of the units that lived longer than $t$, and the simulation is cut off at five times their longest life.
The prediction is then $q_\theta(R_t \mid y^{1:t}, \mathcal{C})$, with $\theta$ unchanged.

\section{Experimental setup}
\label{sec:setup}

We evaluate on nine run-to-failure benchmarks from four domains with published train and test splits: the turbofan subsets C-MAPSS FD001 to FD004~\cite{saxena2008cmapss}, the bearings of PRONOSTIA~\cite{nectoux2012pronostia} and XJTU-SY~\cite{wang2020xjtusy}, the batteries of NB14~\cite{bole2014nb14} and FCLB19~\cite{severson2019fclb19}, and the filters of PHME20~\cite{phme20}, most of them accessed through PHMD~\cite{solismartin2025phmd}.
Each dataset defines its own failure, for example a pressure drop of 20\,psi in PHME20 or 80\,\% of the initial capacity in NB14.

The information taken from the target system is the number $m \in \{1, 2, 4, 8, \text{all}\}$ of run-to-failure training units, drawn per seed and identical for all models.
The baselines are trained on these units: a ridge regression on the raw sensor window as in Picid~\cite{telyatnikov2026picid}, an LSTM~\cite{hochreiter1997lstm}, and the frozen time-series foundation models Timestral (28M parameters)~\cite{timestral2026} and TiRex-2 (83M)~\cite{tirex2}, each with either a ridge head or the two-layer MLP head of El-Ghoussani et al.~\cite{elghoussani2026timeseriesfoundationmodelembeddings}.
With all units, validation units serve for early stopping.
RULix uses the same units only as calibration information (Sec.~\ref{sec:model}) and receives no gradient step.
Each model runs in one fixed configuration for all datasets, without per-dataset tuning, and with three seeds.
The results relate the skill of each model to $m$.

Plain MAE, RMSE or the NASA score depend on the time unit of a dataset and are dominated by its longest units.
We therefore compare models in two steps.
First, we scale the error of each test unit $n \in U$ by its life $T_n$, so that every unit counts equally,
\begin{equation}
  E = \frac{100}{|U|} \sum_{n \in U} \frac{1}{T_n\,|\mathcal{I}_n|} \sum_{t \in \mathcal{I}_n} \big|\hat{R}_{n,t} - R_{n,t}\big| ,
  \label{eq:error}
\end{equation}
where $\mathcal{I}_n$ are the scored steps of unit $n$, $R_{n,t}$ the true and $\hat{R}_{n,t}$ the predicted median RUL.
Second, we relate $E$ to the error $E_{\mathrm{ref}}$ of a lifetime reference,
\begin{equation}
  S = 1 - \frac{E}{E_{\mathrm{ref}}} ,
  \label{eq:skill}
\end{equation}
which reads no sensor and predicts at step $t$ the median residual life of all training units $j$ that lived longer than $t$, $\hat{R}^{\mathrm{ref}}_{n,t} = \operatorname{median}\{T_j - t \mid T_j > t\}$, so a skill of $S = 0$ means no gain over the lifetimes of the training units and $S = 1$ a perfect prediction.
Over all datasets, the skill follows from the geometric mean of $E/E_{\mathrm{ref}}$, with one shared weight for the four C-MAPSS subsets and a 95\,\% bootstrap interval over the test units.
For RULix, we additionally report the coverage of its 80\,\% interval.

\section{Results}
\label{sec:results}

\begin{figure}[t]
\centering
\includegraphics[width=\columnwidth]{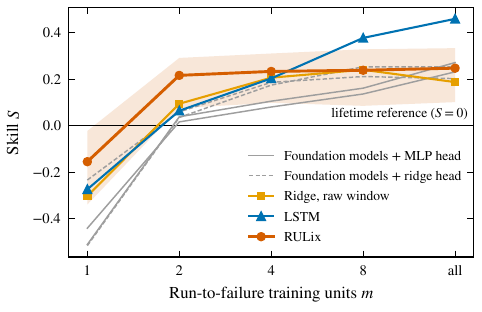}
\caption{Skill $S$~\eqref{eq:skill} over all datasets against the number $m$ of run-to-failure training units. RULix (13M) uses the units only for calibration, all other models are trained on them. Shaded: 95\,\% bootstrap interval of RULix. Grey: frozen Timestral (28M) and TiRex-2 (83M) with MLP head (solid) or ridge head (dashed).}
\label{fig:skill}
\end{figure}

Figure~\ref{fig:skill} shows the skill over the number $m$ of run-to-failure training units.
With up to four units, RULix reaches the highest skill without any training, for example $S = 0.22$ at $m = 2$ against at most $0.09$ for the trained models.
From $m = 8$ on, the LSTM pulls ahead and reaches $S = 0.46$ with all units, while RULix and the foundation models with trained heads stay between $0.20$ and $0.27$.
The negative skills at $m = 1$ arise because the reference uses the lifetimes of all training units, with a single lifetime it reaches $S = -0.60$ itself.
With all units, RULix is better than the reference on every dataset except FCLB19.
Its $80\,\%$ intervals cover $81$ to $92\,\%$ of the true RUL on PHME20 and C-MAPSS FD001, FD002 and FD004, but are too narrow on PRONOSTIA, FCLB19 and XJTU-SY with $22$ to $40\,\%$.

\begin{figure}[t]
\centering
\includegraphics[width=\columnwidth]{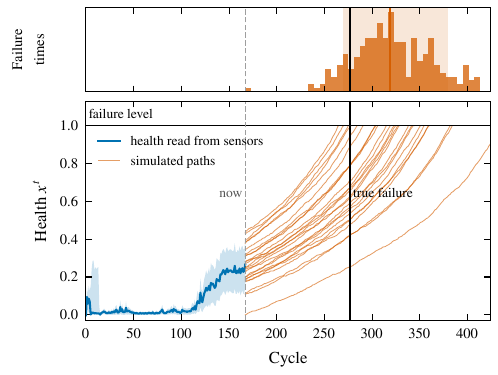}
\caption{One prediction of RULix traced for a held-out C-MAPSS FD001 unit at 60\,\% of its life (dashed). Bottom: health state read from the sensor history up to each cycle and 25 of the 256 paths simulated with the inferred drift and diffusion. Top: failure times of all paths with median and 80\,\% interval, against the true failure (black).}
\label{fig:explain}
\end{figure}

Figure~\ref{fig:explain} traces one prediction.
At $60\,\%$ of the life of a C-MAPSS unit, RULix reads that degradation has begun and starts 256 paths from the current health state.
The paths follow the inferred drift and diffusion until they reach the failure level, and their failure times give a predicted RUL of 152 cycles with an $80\,\%$ interval from 102 to 212 cycles, which contains the true value of 110.
A maintenance planner can thus check each step of the prediction: the health state read from the sensors, how fast degradation is expected to proceed, and how uncertain the resulting failure time is.

\section{Conclusion}
\label{sec:conclusion}

This paper presented RULix, a prognostic foundation model pretrained only on synthetic degradation processes and applied without fine-tuning to nine run-to-failure benchmarks covering turbofans, bearings, batteries and filters.
Calibrated with two to four run-to-failure units, it predicts the RUL more accurately than models trained on the same units, so it can be deployed as soon as the first failures are recorded.
From eight units on, a trained LSTM is more accurate, but it must be retrained for every dataset, whereas RULix keeps its weights and could still be fine-tuned.
Each prediction is traceable through health state, degradation law and simulated failure times.

In future work, the prior could add interacting failure modes, partial repairs and varying sampling rates.
RULix has 13M parameters, fewer than both time-series foundation models, and could be scaled in model size and synthetic data.
Besides fine-tuning, the encoder could take the past units as context and learn the now hand-crafted calibration.
Sampling drift and diffusion per path from their predicted distribution may widen the intervals, and planned instead of constant load would let planners compare maintenance scenarios.

\section*{Acknowledgements}

Parts of the experiments presented in this paper were carried out using the HPC facilities of the University of Luxembourg. This research has been funded by the Federal Ministry of Education and Research of Germany and the state of North Rhine-Westphalia as part of the Lamarr Institute for Machine Learning and Artificial Intelligence.

\bibliographystyle{elsarticle-num}
{\footnotesize\setlength{\bibsep}{2pt}\bibliography{references}}

\begin{thebibliography}{10}
\expandafter\ifx\csname url\endcsname\relax
  \def\url#1{\texttt{#1}}\fi
\expandafter\ifx\csname urlprefix\endcsname\relax\def\urlprefix{URL }\fi
\expandafter\ifx\csname href\endcsname\relax
  \def\href#1#2{#2} \def\path#1{#1}\fi

\bibitem{erkoyuncu2026maintenance}
J.~A. Erkoyuncu, F.~Ansari, B.~Iung, S.~Takata, Digitally enhanced maintenance:
  path towards continuous value creation and sustainability, CIRP Annals 75~(2)
  (2026) 943--969.
\newblock \href {https://doi.org/10.1016/j.cirp.2026.05.002}
  {\path{doi:10.1016/j.cirp.2026.05.002}}.

\bibitem{benker2022maintenance}
M.~Benker, V.~Rommel, M.~F. Zaeh, An investigation into the economic efficiency
  of different maintenance strategies based on a discrete event simulation,
  Procedia CIRP 107 (2022) 428--433.
\newblock \href {https://doi.org/10.1016/j.procir.2022.05.003}
  {\path{doi:10.1016/j.procir.2022.05.003}}.

\bibitem{kaiser2023sme}
J.~Kaiser, G.~Terrazas, D.~McFarlane, L.~de~Silva, Towards low-cost machine
  learning solutions for manufacturing {SMEs}, AI \& Society 38~(6) (2023)
  2659--2665.
\newblock \href {https://doi.org/10.1007/s00146-021-01332-8}
  {\path{doi:10.1007/s00146-021-01332-8}}.

\bibitem{welte2020sme}
R.~Welte, M.~Estler, D.~Lucke, A method for implementation of machine learning
  solutions for predictive maintenance in small and medium sized enterprises,
  Procedia CIRP 93 (2020) 909--914.
\newblock \href {https://doi.org/10.1016/j.procir.2020.04.052}
  {\path{doi:10.1016/j.procir.2020.04.052}}.

\bibitem{li2024smalldata}
C.~Li, S.~Li, Y.~Feng, K.~Gryllias, F.~Gu, et~al., Small data challenges for
  intelligent prognostics and health management: a review, Artificial
  Intelligence Review 57~(8) (2024) 214.
\newblock \href {https://doi.org/10.1007/s10462-024-10820-4}
  {\path{doi:10.1007/s10462-024-10820-4}}.

\bibitem{xiong2023physics}
J.~Xiong, O.~Fink, J.~Zhou, Y.~Ma, Controlled physics-informed data generation
  for deep learning-based remaining useful life prediction under unseen
  operation conditions, Mechanical Systems and Signal Processing 197 (2023)
  110359.
\newblock \href {https://doi.org/10.1016/j.ymssp.2023.110359}
  {\path{doi:10.1016/j.ymssp.2023.110359}}.

\bibitem{kim2025sparse}
E.~Kim, H.~Lee, Y.~Sim, J.~Lee, M.~B.~G. Jun, Overcoming sparse run-to-failure
  data challenges in manufacturing: A contrastive mixer framework for remaining
  useful life prediction, CIRP Annals 74~(1) (2025) 579--583.
\newblock \href {https://doi.org/10.1016/j.cirp.2025.04.082}
  {\path{doi:10.1016/j.cirp.2025.04.082}}.

\bibitem{theiler2026tabpfn}
R.~Theiler, L.~Telyatnikov, L.~von Krannichfeldt, O.~Fink, Towards unified and
  data-efficient prognostics and health management with tabular foundation
  models, arXiv preprint (2026).
\newblock \href {https://doi.org/10.48550/arXiv.2606.05481}
  {\path{doi:10.48550/arXiv.2606.05481}}.

\bibitem{nunes2022challenges}
P.~Nunes, J.~Santos, E.~Rocha, Challenges in predictive maintenance – a
  review, CIRP Journal of Manufacturing Science and Technology 40 (2023)
  53--67.
\newblock \href {https://doi.org/10.1016/j.cirpj.2022.11.004}
  {\path{doi:10.1016/j.cirpj.2022.11.004}}.

\bibitem{braun2026piml}
C.~Braun, J.~Raible, M.~F. Huber, Physics-informed machine learning in
  prognostics and health management: a systematic literature review, Journal of
  Intelligent Manufacturing (2026).
\newblock \href {https://doi.org/10.1007/s10845-026-02930-3}
  {\path{doi:10.1007/s10845-026-02930-3}}.

\bibitem{pauly2026pumps}
A.~Pauly, R.~Fischer, N.~Jamialahmadi, D.~Graurock, M.~Oettmeier, et~al.,
  Retrofitting water pumps with health monitoring and machine learning at the
  edge, Procedia CIRP 146 (2026) 672--677.
\newblock \href {https://doi.org/10.1016/j.procir.2026.03.280}
  {\path{doi:10.1016/j.procir.2026.03.280}}.

\bibitem{elghoussani2026timeseriesfoundationmodelembeddings}
A.~El-Ghoussani, M.~De~Vita, R.~Naumann, V.~Belagiannis, Time-series foundation
  model embeddings for remaining useful life estimation, arXiv preprint (2026).
\newblock \href {https://doi.org/10.48550/arXiv.2606.11990}
  {\path{doi:10.48550/arXiv.2606.11990}}.

\bibitem{jardine_review_2006}
A.~K.~S. Jardine, D.~Lin, D.~Banjevic, A review on machinery diagnostics and
  prognostics implementing condition-based maintenance, Mechanical Systems and
  Signal Processing 20~(7) (2006) 1483--1510.
\newblock \href {https://doi.org/10.1016/j.ymssp.2005.09.012}
  {\path{doi:10.1016/j.ymssp.2005.09.012}}.

\bibitem{si2010rulstat}
X.-S. Si, W.~Wang, C.-H. Hu, D.-H. Zhou, Remaining useful life estimation – a
  review on the statistical data driven approaches, European Journal of
  Operational Research 213~(1) (2011) 1--14.
\newblock \href {https://doi.org/10.1016/j.ejor.2010.11.018}
  {\path{doi:10.1016/j.ejor.2010.11.018}}.

\bibitem{qi2026survivalpfn}
S.~Qi, V.~Balazadeh, M.~Cooper, R.~Greiner, R.~G. Krishnan, {SurvivalPFN}:
  Amortizing survival prediction via in-context {B}ayesian inference, arXiv
  preprint (2026).
\newblock \href {https://doi.org/10.48550/arXiv.2605.15488}
  {\path{doi:10.48550/arXiv.2605.15488}}.

\bibitem{sarkka2019applied}
S.~Särkkä, A.~Solin, Applied Stochastic Differential Equations, Cambridge
  University Press, 2019.
\newblock \href {https://doi.org/10.1017/9781108186735}
  {\path{doi:10.1017/9781108186735}}.

\bibitem{mueller2021pfn}
S.~M{\"u}ller, N.~Hollmann, S.~Pineda~Arango, J.~Grabocka, F.~Hutter,
  \href{https://openreview.net/forum?id=KSugKcbNf9}{Transformers can do
  {B}ayesian inference}, in: International Conference on Learning
  Representations (ICLR 2022), 2022.
\newline\urlprefix\url{https://openreview.net/forum?id=KSugKcbNf9}

\bibitem{Gershman2014AmortizedII}
S.~Gershman, N.~D. Goodman,
  \href{https://escholarship.org/uc/item/34j1h7k5}{Amortized inference in
  probabilistic reasoning}, in: Proceedings of the 36th Annual Meeting of the
  Cognitive Science Society (CogSci 2014), 2014.
\newline\urlprefix\url{https://escholarship.org/uc/item/34j1h7k5}

\bibitem{seifner2025incontext}
P.~Seifner, K.~Cvejoski, D.~Berghaus, C.~Ojeda, R.~J. Sanchez,
  \href{https://openreview.net/forum?id=ceCJPoZOKJ}{In-context learning of
  stochastic differential equations with foundation inference models}, in:
  Advances in Neural Information Processing Systems 38 (NeurIPS 2025), 2025.
\newline\urlprefix\url{https://openreview.net/forum?id=ceCJPoZOKJ}

\bibitem{beck2024xlstm}
M.~Beck, K.~P{\"{o}}ppel, M.~Spanring, A.~Auer, O.~Prudnikova, et~al., {xLSTM}:
  Extended long short-term memory, in: Advances in Neural Information
  Processing Systems 37 (NeurIPS 2024), 2024.
\newblock \href {https://doi.org/10.52202/079017-3417}
  {\path{doi:10.52202/079017-3417}}.

\bibitem{saxena2008cmapss}
A.~Saxena, K.~Goebel, D.~Simon, N.~Eklund, Damage propagation modeling for
  aircraft engine run-to-failure simulation, in: IEEE International Conference
  on Prognostics and Health Management (PHM 2008), 2008, pp. 1--9.
\newblock \href {https://doi.org/10.1109/PHM.2008.4711414}
  {\path{doi:10.1109/PHM.2008.4711414}}.

\bibitem{nectoux2012pronostia}
P.~Nectoux, R.~Gouriveau, K.~Medjaher, E.~Ramasso, B.~Chebel-Morello, et~al.,
  \href{https://hal.science/hal-00719503}{{PRONOSTIA: An experimental platform
  for bearings accelerated degradation tests}}, in: IEEE International
  Conference on Prognostics and Health Management (PHM 2012), 2012, pp. 1--8.
\newline\urlprefix\url{https://hal.science/hal-00719503}

\bibitem{wang2020xjtusy}
B.~Wang, Y.~Lei, N.~Li, N.~Li, A hybrid prognostics approach for estimating
  remaining useful life of rolling element bearings, IEEE Transactions on
  Reliability 69~(1) (2020) 401--412.
\newblock \href {https://doi.org/10.1109/TR.2018.2882682}
  {\path{doi:10.1109/TR.2018.2882682}}.

\bibitem{bole2014nb14}
B.~Bole, C.~S. Kulkarni, M.~Daigle, Adaptation of an electrochemistry-based
  {Li-Ion} battery model to account for deterioration observed under randomized
  use, Annual Conference of the PHM Society 6~(1) (2014).
\newblock \href {https://doi.org/10.36001/phmconf.2014.v6i1.2490}
  {\path{doi:10.36001/phmconf.2014.v6i1.2490}}.

\bibitem{severson2019fclb19}
K.~A. Severson, P.~M. Attia, N.~Jin, N.~Perkins, B.~Jiang, et~al., Data-driven
  prediction of battery cycle life before capacity degradation, Nature Energy
  4~(5) (2019) 383--391.
\newblock \href {https://doi.org/10.1038/s41560-019-0356-8}
  {\path{doi:10.1038/s41560-019-0356-8}}.

\bibitem{phme20}
K.~Ince, E.~Sirkeci, Y.~Genç, Remaining useful life prediction for
  experimental filtration system: A data challenge, PHM Society European
  Conference 5~(1) (2020).
\newblock \href {https://doi.org/10.36001/phme.2020.v5i1.1317}
  {\path{doi:10.36001/phme.2020.v5i1.1317}}.

\bibitem{solismartin2025phmd}
D.~Solís-Martín, J.~Galán-Páez, J.~Borrego-Díaz, {PHMD}: An easy data
  access tool for prognosis and health management datasets, SoftwareX 29 (2025)
  102039.
\newblock \href {https://doi.org/10.1016/j.softx.2025.102039}
  {\path{doi:10.1016/j.softx.2025.102039}}.

\bibitem{telyatnikov2026picid}
L.~Telyatnikov, R.~Theiler, L.~von Krannichfeldt, O.~Fink, {Picid}: A modular
  evaluation infrastructure for reproducible {PHM} across tasks and domains,
  arXiv preprint (2026).
\newblock \href {https://doi.org/10.48550/arXiv.2605.28345}
  {\path{doi:10.48550/arXiv.2605.28345}}.

\bibitem{hochreiter1997lstm}
S.~Hochreiter, J.~Schmidhuber, Long short-term memory, Neural Computation 9~(8)
  (1997) 1735--1780.
\newblock \href {https://doi.org/10.1162/neco.1997.9.8.1735}
  {\path{doi:10.1162/neco.1997.9.8.1735}}.

\bibitem{timestral2026}
A.~Petralia, A.~Gavaudan, A.~Joosen, E.~Goffinet, R.~Kurle, et~al., Timestral,
  arXiv preprint (2026).
\newblock \href {https://doi.org/10.48550/arXiv.2610.37585}
  {\path{doi:10.48550/arXiv.2610.37585}}.

\bibitem{tirex2}
P.~Podest, M.~Pichler, E.~Bürger, L.~Zólyomi, B.~Voggenberger, et~al.,
  {TiRex-2}: Generalizing {TiRex} to multivariate data and streaming, arXiv
  preprint (2026).
\newblock \href {https://doi.org/10.48550/arXiv.2607.01204}
  {\path{doi:10.48550/arXiv.2607.01204}}.

\end{thebibliography}

\end{document}